\documentclass[runningheads]{llncs}
\usepackage[T1]{fontenc}
\usepackage{graphicx,verbatim}
\usepackage{amsfonts}
\usepackage{amsmath}
\usepackage{bm}
\usepackage{multirow}
\usepackage{xcolor}
\usepackage{boldline}
\usepackage{marvosym}
\begin{document}
\title{AdvMIM: Adversarial Masked Image Modeling for Semi-Supervised Medical Image Segmentation}
\titlerunning{AdvMIM for Semi-Supervised Medical Image Segmentation}

\author{Lei Zhu\inst{1\text{\Letter}} \and
Jun Zhou\inst{1} \and
Rick Siow Mong Goh\inst{1}\and
Yong Liu\inst{1}}
\authorrunning{Lei Zhu et al.}
\institute{Institute of High Performance Computing (IHPC), Agency for Science, Technology and Research (A*STAR), 1 Fusionopolis Way, \#16-16 Connexis, Singapore 138632, Republic of Singapore \\ \email{zhu\_lei@ihpc.a-star.edu.sg}}
    
\maketitle              
\begin{abstract}
Vision Transformer (ViT) has recently gained tremendous popularity in medical image segmentation task due to its superior capability in capturing long-range dependencies. However, transformer requires a large amount of labeled data to be effective, which hinders its applicability in annotation scarce semi-supervised learning scenario where only limited labeled data is available. State-of-the-art semi-supervised learning methods propose combinatorial CNN-Transformer learning to cross teach a transformer with a convolutional neural network (CNN), which achieves promising results. However, it remains a challenging task to effectively train the transformer with limited labeled data. In this paper, we propose an adversarial masked image modeling (\textbf{AdvMIM}) method to fully unleash the potential of transformer for semi-supervised medical image segmentation. The key challenge in semi-supervised learning with transformer lies in the lack of sufficient supervision signal. To this end, we propose to construct an auxiliary masked domain from original domain with masked image modeling and train the transformer to predict the entire segmentation mask with masked inputs to increase supervision signal. We leverage the original labels from labeled data and pseudo-labels from unlabeled data to learn the masked domain. To further benefit the original domain from masked domain, we provide a theoretical analysis of our method from a multi-domain learning perspective and devise a novel adversarial training loss to reduce the domain gap between the original and masked domain, which boosts semi-supervised learning performance. We also extend adversarial masked image modeling to CNN network. Extensive experiments on three public medical image segmentation datasets demonstrate the effectiveness of our method, where our method outperforms existing methods significantly. Our code is publicly available at https://github.com/zlheui/AdvMIM.


\end{abstract}

\section{Introduction}
Medical image segmentation is an important task for computer assisted diagnosis, treatment planning, and intervention. With the recent advancement of vision transformer~\cite{dosovitskiy2020image} and its exceptional capability in capturing long-range dependencies, there is growing interest in the medical domain to leverage transformer for medical image segmentation task~\cite{chen2021transunet,cao2022swin}. However, vision transformer is even more annotation hungry than convolutional neural network (CNN)~\cite{touvron2020training}. Semi-supervised learning methods aim to leverage a large amount of unlabeled data together with a limited amount of labeled data for learning a segmentation network to reduce the annotation cost. Existing semi-supervised learning methods can be categorized into pseudo-labeling based methods~\cite{bai2017semi,vu2019advent,chen2021semi}, consistency regularization based methods~\cite{tarvainen2017mean,yu2019uncertainty,ouali2020semi,luo2021efficient,verma2022interpolation,basak2022exceedingly}, deep co-training based methods~\cite{qiao2018deep,chen2021semi,zhu2021semi}, and adversarial training based methods~\cite{zhang2017deep,hu2020coarse}. However, benchmarking results~\cite{luo2022semi,huang2024combinatorial} indicate that directly integrating these methods with transformer leads to poor performance, likely due to the annotation dependency of transformer. State-of-the-art semi-supervised learning method~\cite{luo2022semi} proposes cross-teaching to leverage the complementary architectural advantages of both CNN with efficient local features learning and transformer with better long-range dependencies capturing for the task. Most recently, Huang et al.~\cite{huang2024combinatorial} propose a combinatorial CNN-Transformer learning framework at manifold space with intra-student consistency regularization and inter-student knowledge transfer, which achieves state-of-the-art performance on multiple datasets. While these methods have achieved promising results, it remains a challenging task to effectively train the transformer with limited labeled data.

In this paper, we propose an adversarial masked image modeling (\textbf{AdvMIM}) method to fully unleash the potential of transformer for semi-supervised medical image segmentation. The key challenge in semi-supervised learning with transformer lies in the lack of sufficient supervision signal. While combinatorial CNN-Transformer learning based methods~\cite{luo2022semi,huang2024combinatorial} leverage a CNN network to assign pseudo-labels to unlabeled data for training the transformer, which boosts model performance, we believe even more supervision signal is needed to effectively train the transformer. Therefore, we propose to construct an auxiliary masked domain from original domain with masked image modeling~\cite{xie2022simmim} and perform masked domain learning with transformer to increase the supervision signal. Specifically, masked image modeling~\cite{xie2022simmim} is an effective self-supervised learning method for vision transformer, where the task is to reconstruct the masked image patches. We employ the same masking operation to construct a masked domain and train the vision transformer to predict the entire segmentation mask from the masked inputs. We utilize the original labels from labeled data and pseudo-labels from unlabeled data to learn the masked domain. With the new input and new task, the transformer gains extra supervision signal for learning. To further benefit the original domain from masked domain, we provide a theoretical analysis of our framework from a multi-domain learning perspective and devise a novel adversarial training loss to reduce the domain gap between the original and masked domain, where we employ a domain discriminator to distinguish the prediction masks of both original labeled data and masked unlabeled data and adversarially train the transformer to produce more accurate prediction masks for the unlabeled masked data so that the discriminator cannot distinguish them from those of labeled data.

In summary, we have made the following contributions in this paper: \textbf{(1).} We propose an adversarial masked image modeling method to fully unleash the potential of transformer for semi-supervised medical image segmentation; \textbf{(2).} We provide a theoretical analysis of our method from a multi-domain learning perspective and propose a novel adversarial training loss to reduce the domain gap between masked and original domain; \textbf{(3).} We extend adversarial masked image modeling to CNN network; \textbf{(4).} We perform extensive experiments to evaluate our method on three public medical image segmentation datasets, where our method outperforms existing methods significantly.

\begin{figure}[t]
\centering
\includegraphics[width=0.5\textwidth]{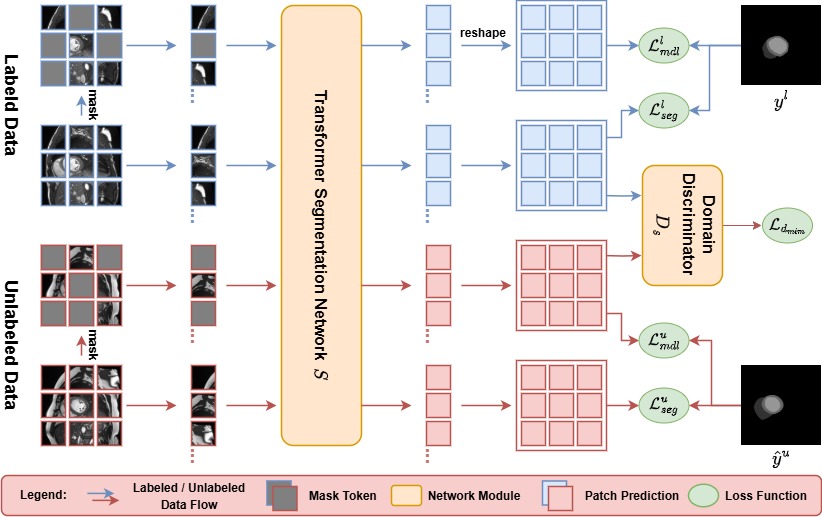}
\caption{Architecture and dataflow of our proposed adversarial masked image modeling method. Our method constructs an auxiliary masked domain from original domain with masked image modeling. We utilize original labels from labeled data and pseudo-labels from unlabeled data to learn the masked domain. We further propose a novel adversarial training loss to reduce the domain gap between the original and masked domain to boost semi-supervised learning performance. Note pseudo-label $\hat{y}$ is obtained from a cross-teaching CNN.}
\label{fig1}
\end{figure}

\section{Methodology}
In semi-supervised medical image segmentation, we are given $N^l$ labeled data $\mathbb{D}^{l}=\{(x_{i}^l,y_{i}^l)\}^{N^l}_{i=1}$ and $N^u$ unlabeled data $\mathbb{D}^{u}=\{x_{i}^u\}^{N^u}_{i=1}$, where $y_{i}^l$ is the corresponding segmentation mask for $x_{i}^l$ with $M$ classes and $N^l << N^u$. Both the labeled data and unlabeled data are sampled from probability distribution $P$, where unlabeled data are sampled without labels. The goal is to learn an accurate segmentation model with both labeled and unlabeled data. Fig.~\ref{fig1} presents an overview of our proposed adversarial masked image modeling method.

\subsection{Segmentation}
As illustrated in Fig.~\ref{fig1}, we train the transformer segmentation network $S:\mathbb{R}^{H\times W\times 3}\rightarrow\mathbb{R}^{H\times W\times M}$ with both labeled data and pseudo-labeled unlabeled data. We follow existing methods~\cite{luo2022semi,huang2024combinatorial} to cross teach the transformer network with a convolutional neural network $C:\mathbb{R}^{H\times W\times 3}\rightarrow\mathbb{R}^{H\times W\times M}$, where the pseudo-label for unlabeled data is obtained through $\hat{y}_i^u=argmax(C(x_i^u))$. As pseudo-label may contain noise, we weight the loss function with its maximum predicted probability as a certainty measure~\cite{lee2013pseudo}, where the weight is calculated as $w_i^u=max(C(x_i^u))$. The segmentation loss functions on labeled and unlabeled data for the transformer are defined as follows:
\begin{equation} \label{eqn:0} 
\mathcal{L}_{seg}^l(S)=\mathbb{E}_{x^l\sim \mathbb{D}^l} [H(y^l,S(x^l)) + Dice(y^l, S(x^l))], 
\end{equation}
\begin{equation}\label{eqn:1} 
\mathcal{L}_{seg}^u(S)=\mathbb{E}_{x^u\sim \mathbb{D}^u} [w^uH(\hat{y}^u,S(x^u)) + Dice(\hat{y}^u, S(x^u))], 
\end{equation}
where $H(\cdot)$ calculates the pixel-wise cross-entropy loss and $Dice(\cdot)$ calculates the \textit{Dice} loss. Note the CNN is trained together with the transformer using the same loss functions except that pseudo-labels for CNN are assigned by transformer.

\subsection{Masked Domain Learning}
The key challenge in semi-supervised medical image segmentation with vision transformer lies in the lack of sufficient supervision signal. We propose to construct an auxiliary masked domain from the original domain with masked image modeling~\cite{xie2022simmim}. But instead of reconstructing the masked image patches, we propose to train the transformer to predict the entire segmentation mask with masked input, where the transformer learns to infer the semantics on masked image patches based on the visible ones. We employ the same masking operation in masked image modeling to construct masked images. We replace the masked image patches with shared learnable mask tokens with positional embedding and utilize original labels from labeled data and pseudo-labels from unlabeled data to learn the masked domain. With the new input and new task, the transformer gains extra supervision signals for learning. The masked domain learning loss for labeled and unlabeled data are defined as follows:
\begin{equation} \label{eqn:2}
\mathcal{L}_{mdl}^l(S)=\mathbb{E}_{x^l\sim \mathbb{D}^l} [H(y^l,S(x^{ml}))) + Dice(y^l, S(x^{ml})], 
\end{equation}
\begin{equation} \label{eqn:3} 
\mathcal{L}_{mdl}^u(S)=\mathbb{E}_{x^u\sim \mathbb{D}^u} [w^uH(\hat{y}^u,S(x^{mu}) + Dice(\hat{y}^u, S(x^{mu})], 
\end{equation}
where $x^{ml} = [\mathcal{M}(x^l,\rho);\bm{T}]$ is the masked labeled data, $\mathcal{M}(\cdot,\rho)$ denotes the masking operation with mask ratio $\rho$, $\bm{T}$ denotes the set of mask tokens to replace the masked image patches with positional embedding, the operation $[\cdot;\cdot]$ concatenates two input vectors into a single vector, and $x^{mu} = [\mathcal{M}(x^u,\rho);\bm{T}]$. Following~\cite{xie2022simmim}, we set $\rho=0.7$ by default.

\subsection{Adversarial Masked Domain Adaptation}
We treat both the pseudo-labeled original domain and the pseudo-labeled masked domain as noisily labeled domain. Inspired by ~\cite{ben2010theory}, we analyze our framework from a multi-domain learning perspective and present the following theorem:

\begin{theorem} [Masked Domain Adaptation Theorem] \label{theorem1} Following the problem definition, let $h$ be a hypothesis in class $\mathcal{H}$, let $\gamma$ be the pseudo-label noise ratio, denote the probability distribution of pseudo-labeled original domain as $P'$ and the probability distribution of pseudo-labeled masked domain as $Q'$, then for any $\delta \in (0,1)$, with probability at least $1-\delta$, for every $h \in \mathcal{H}$, we have:
\begin{equation}
\label{eqn:4}
\begin{split}
\epsilon_P(h) \leq & \frac{1}{2}\epsilon_{P'}(h) + \frac{1}{2}\epsilon_{Q'}(h) +\frac{1}{4}d_{\mathcal{H}\Delta\mathcal{H}}(P, Q') + \frac{1}{2}\lambda + \gamma,
\end{split}
\end{equation}
\end{theorem}
where $\epsilon_P(\cdot)$ (resp. $\epsilon_{P'}(\cdot)$, $\epsilon_{Q'}(\cdot)$) measures the expectation error of a hypothesis on original (resp. pseudo-labeled original, pseudo-labeled masked) data distribution, $d_{\mathcal{H}\Delta\mathcal{H}}(\cdot, \cdot)$ measures the distribution discrepancy between two data distributions, and $\lambda=\min_{h \in \mathcal{H}} \epsilon_{P}(h)+\epsilon_{Q}(h)$.
\begin{proof} [Sketch]
We utilize the triangle inequality to bound the expectation error with the pseudo-label noisy ratio. We leverage $Lemma\;4$ in~\cite{ben2010theory} to relate the expectation error on original domain with multi-domain learning loss on original and pseudo-labeled masked domain. With equation substitution, we refine the bound to explicitly capture the pseudo-label noise ratio to obtain the final bound.
\end{proof}
The theorem upper bounds expectation error on original domain with (1). expectation error on pseudo-labeled original domain; (2). expectation error on pseudo-labeled masked domain; (3). the domain gap between original and masked domain; (4). pseudo-label noisy ratio; and (5). the non-optimizable optimal error between original and pseudo-labeled masked domain that is assumed to be small~\cite{ben2010theory}. Our segmentation loss and masked domain learning loss functions minimize term (1) and term (2) respectively. We weight the loss functions with certainty measures to minimize term (4). Except for the non-optimizable term (5), our theorem further indicates that it is necessary to minimize term (3), the domain gap between original and masked domain to fully bound the expectation error on original domain. 

To this end, we introduce a domain discriminator $D_s: \mathbb{R}^{H\times W\times M}\rightarrow\mathbb{R}$, which takes the prediction masks from both original labeled data and masked unlabeled data as input and outputs the domain prediction. We adversarially train the transformer to confuse the domain discriminator, where the transformer and the domain discriminator play a two-player min-max game following the GAN~\cite{goodfellow2014generative} framework. At the optimal, the transformer aligns the masked unlabeled data distribution towards the original labeled data distribution to reduce the domain gap between masked and original domain so that the discriminator cannot distinguish the prediction masks between them anymore. Following~\cite{mao2017least}, we adopt the least squares loss for GAN training to enhance the training tability. The adversarial masked image modeling loss is defined as follow:
\begin{equation} \label{eqn:5}
    \begin{split}
    \mathcal{L}_{d_{mim}}(D_s) &= \mathbb{E}_{x^l\sim \mathbb{D}^l} [(D_s(S(x^l))-1)^2] \\&+ \mathbb{E}_{x^u\sim \mathbb{D}^u} [(D_s(S(x^{mu})))^2], \\
    \end{split}
\end{equation}
\begin{equation} \label{eqn:6}
    \begin{split}
    \mathcal{L}_{adv_{mim}}(S) = \mathbb{E}_{x^u\sim \mathbb{D}^u} [(D_s(S(x^{mu}))-1)^2]. \\
    \end{split}
\end{equation}

\noindent\textbf{Discussion.} Note different from existing adversarial training based semi-supervised learning method~\cite{zhang2017deep}, which performs adversarial training to reduce the domain gap between labeled and unlabeled data, our method constructs and learns an auxiliary masked domain and performs adversarial training to reduce the domain gap between original and masked domain on labeled and masked unlabeled data.

\subsection{Extension to CNN}
We propose to extend adversarial masked image modeling to CNN network to improve its learning with limited labeled data, which in turn can benefit the learning of transformer through the cross-teaching process. Specifically, we apply the same masking operation and perform masked domain learning and adversarial masked domain adaptation on CNN network. We add a domain discriminator $D_{c}$ to reduce the domain gap between original and masked domain in CNN branch. Due to the symmetry of transformer and CNN branch, the same loss functions are defined for CNN network. 
Without loss of generality, the overall objective of our framework is defined as follows:
\begin{equation} \label{eqn:7}
    \begin{split}
    \min_{S,C}\;&\mathcal{L}_{seg}^l + \mathcal{L}_{seg}^u + \mathcal{L}_{mdl}^l + \mathcal{L}_{mdl}^u + \lambda_{adv} \mathcal{L}_{adv_{mim}}, \\
    \min_{D_s,D_c}\;&\mathcal{L}_{d_{mim}},
    \end{split}
\end{equation}
where $\lambda_{adv}$ is balancing weight which is empirically set to be 0.001.

\section{Experimental Analysis} 
\noindent\textbf{Datasets.} We evaluate the effectiveness of our framework on three public datasets, namely Automated Cardiac Diagnosis Challenge (ACDC)~\cite{bernard2018deep}, Synapse~\cite{landman2015segmentation}, and International Skin Imaging Collaboration (ISIC)~\cite{codella2018skin}. ACDC contains 100 magnetic resonance imaging (MRI) scans of three organs. Following~\cite{luo2022semi}, we adopt 70, 10 and 20 cases for training, validation and testing. We evaluate with 3\% and 10\% partitions of training data as labeled data, while the rest training data as unlabeled data for semi-supervised segmentation. Synapse consists of 30 computed tomography (CT) scans annotated with eight abdominal organs. Following~\cite{huang2024combinatorial}, we adopt 18 and 12 cases for training and testing and evaluate with 15\% and 30\% partitions of the training data. ISIC is a skin lesion segmentation dataset including 2,594 dermoscopy images, with 1,838 training images and 756 validation images. We experiment with 3\% and 10\% partitions of training data.

\noindent\textbf{Implementation.} In all experiments, we adopt Swin-UNet~\cite{cao2022swin} as the transformer segmentation network and UNet~\cite{ronneberger2015u} as the convolutional neural network. The UNet is only used for complementary training and not used for final prediction. We implement the domain discriminator with a five-layer CNN network. We train our framework with SGD optimizer for 30,000 iterations, where the initial learning rate is 0.05, momentum is 0.9 and weight decay is 1e-4. The batch size is 16 with half labeled and half unlabeled images. Following~\cite{huang2024combinatorial}, we randomly crop a patch with size of 224$\times$224 as the input. We perform standard data augmentation to avoid overfitting, including random flip and rotation. We employ two commonly-used metrics, the Dice coefficient (Dice) and the Hausdorff Distance (HD) to quantitatively evaluate the segmentation performance.

\begin{table}[t]
\caption{Ablation study on ACDC (10\%) and Synapse (15\%) in Dice (\%) and HD (mm). The best performance is marked in \textbf{bold}.} 
\label{table:1}
\centering
\begin{small}
\resizebox{0.7\linewidth}{!}{%
\begin{tabular}{c|cc|cc|cc|cc|>{\centering\arraybackslash}p{0.5cm}c|cccc}
\hlineB{3}
\multirow{2}{*}{Method}&\multicolumn{2}{c|}{$\mathcal{L}_{seg}^l$} & \multicolumn{2}{c|}{$\mathcal{L}_{seg}^u$} & \multicolumn{2}{c|}{$\mathcal{L}_{mdl}^l$} & \multicolumn{2}{c|}{$\mathcal{L}_{mdl}^u$} & \multicolumn{2}{c|}{$\mathcal{L}_{adv_{mim}}$} & \multicolumn{2}{c}{ACDC (10\%)}  & \multicolumn{2}{c}{Synapse (15\%)} \\
\cline{2-15}
&S&C&S&C&S&C&S&C&S&C&Dice&HD&Dice&HD\\
\hline
Labeled only&\checkmark &  &  &  &  &  &  &  &  &  & 79.6 & 4.1 & 45.7 & 43.1\\
+Cross Teaching\textsubscript{(weighted)}&\checkmark & \checkmark & \checkmark & \checkmark &  &  &  &  &  &  & 86.6 & 2.5 & 63.1 & 26.7\\
+Masked Domain Learning&\checkmark & \checkmark & \checkmark & \checkmark & \checkmark & & \checkmark &  &  &  & 88.2 & 1.8 & 65.0 & 23.5\\
+Adversarial Masked Domain Adaptation&\checkmark & \checkmark & \checkmark & \checkmark & \checkmark &  & \checkmark &  & \checkmark &  & 88.9 & \textbf{1.3} & 65.8 & 22.9\\
\textbf{AdvMIM}&\checkmark & \checkmark & \checkmark & \checkmark & \checkmark & \checkmark & \checkmark & \checkmark & \checkmark & \checkmark & \textbf{89.0} & \textbf{1.3} & \textbf{66.3} & \textbf{22.7}\\
\hlineB{3}
\end{tabular}}
\end{small}
\end{table}

\begin{table*}[t]
\caption{Comparison with SoTA methods on ACDC, Synapse, and ISIC in Dice (\%) and HD (mm). The best results are in \textbf{bold}, and the second-best results are \underline{underlined}.} 
\label{table:sota_comparison}
\centering
\resizebox{0.95\linewidth}{!}{%
\begin{tabular}{l| cc cc |cc cc| cc cc}
\hlineB{3}
\multirow{2}{*}{Method} 
& \multicolumn{2}{c}{ACDC (3\%)} & \multicolumn{2}{c|}{ACDC (10\%)} & \multicolumn{2}{c}{Synapse (15\%)} & \multicolumn{2}{c|}{Synapse (30\%)} & \multicolumn{2}{c}{ISIC (3\%)} 
& \multicolumn{2}{c}{ISIC (10\%)} \\
\cline{2-13}
& Dice & HD & Dice & HD & Dice & HD & Dice & HD & Dice & HD & Dice & HD \\
\hline
MT~\cite{tarvainen2017mean} & 56.6 & 34.5 & 81.0 & 14.4 & 49.7 & 69.4 & 61.1 & 63.8 & 72.8 & 37.4 & 73.4 & 34.0 \\
UA-MT~\cite{yu2019uncertainty} & 61.0 & 25.8 & 81.5 & 14.4 & 51.3 & 93.4 & 57.8 & 63.9 & 73.0 & 38.6 & 73.4 & 33.2 \\
EM~\cite{vu2019advent} & 60.2 & 24.1 & 79.1 & 14.5 & 49.5 & 72.7 & 59.7 & 63.8 & 72.3 & 36.3 & 72.7 & 39.3 \\
DCT~\cite{qiao2018deep} & 58.2 & 26.4 & 80.4 & 13.8 & 51.0 & 77.0 & 60.6 & 64.2 & 72.9 & 40.6 & 76.0 & 35.7 \\
CCT~\cite{ouali2020semi} & 58.6 & 27.9 & 81.6 & 13.1 & 40.2 & 75.9 & 57.6 & 69.9 & 67.7 & 42.2 & 72.3 & 31.7 \\
CPS~\cite{chen2021semi} & 60.3 & 25.5 & 83.3 & 11.0 & 47.9 & 66.2 & 60.7 & 69.0 & 68.6 & 44.4 & 74.3 & 35.7 \\
ICT~\cite{verma2022interpolation} & 58.1 & 22.8 & 81.1 & 11.4 & 52.7 & 70.5 & 62.7 & 59.6 & 73.2 & 37.2 & 75.3 & 34.6 \\
DAN~\cite{zhang2017deep} & 52.8 & 32.6 & 79.5 & 14.6 & 47.0 & 93.3 & 58.3 & 73.3 & 69.5 & 39.5 & 72.4 & 30.4 \\
URPC~\cite{luo2021efficient} & 56.7 & 31.4 & 82.9 & 10.6 & 48.9 & 69.6 & 59.7 & 66.0 & 70.3 & 39.3 & 75.8 & 32.8 \\
CTCT~\cite{luo2022semi} & 70.4 & 12.4 & 86.4 & 8.6 & 60.4 & 45.4 & 68.7 & 44.3 & 71.3 & 43.2 & 76.0 & 37.3 \\
SSNet~\cite{wu2022exploring} & 70.5 & 17.4 & 85.3 & 10.6 & 58.1 & 47.3 & 66.8 & 34.9 & 72.8 & 40.8 & 75.8 & 32.8 \\
ICT-Med~\cite{basak2022exceedingly} & 56.3 & 22.6 & 83.7 & 13.1 & 51.5 & 62.0 & 61.2 & 59.1 & 71.4 & 39.2 & 74.9 & 33.1 \\
M-CnT~\cite{huang2024combinatorial} & \underline{75.3} & \underline{10.7} & \underline{88.4} & \underline{4.4} & \underline{65.3} & \underline{32.6} & \underline{71.4} & \underline{31.2} & \underline{77.9} & \underline{32.1} & \underline{81.1} & \underline{24.4} \\
\hline
\textbf{AdvMIM} & $\textbf{85.4}_{(10.1\textcolor{green}{\uparrow})}$ & $\textbf{2.0}_{(8.7\textcolor{red}{\downarrow})}$ & $\textbf{89.0}_{(0.6\textcolor{green}{\uparrow})}$ & $\textbf{1.3}_{(3.1\textcolor{red}{\downarrow})}$ & $\textbf{66.3}_{(1.0\textcolor{green}{\uparrow})}$ & $\textbf{22.7}_{(9.9\textcolor{red}{\downarrow})}$ & $\textbf{74.8}_{(3.4\textcolor{green}{\uparrow})}$ & $\textbf{15.5}_{(15.7\textcolor{red}{\downarrow})}$ & $\textbf{79.8}_{(1.9\textcolor{green}{\uparrow})}$ & $\textbf{22.2}_{(9.9\textcolor{red}{\downarrow})}$ & $\textbf{81.8}_{(0.7\textcolor{green}{\uparrow})}$ & $\textbf{19.5}_{(4.9\textcolor{red}{\downarrow})}$ \\
\hlineB{3}
\end{tabular}}
\end{table*}

\noindent\textbf{Ablation Study.} In Table~\ref{table:1}, we present the ablation study of different components of our method. As can be observed, the baseline labeled only method performs poorly. 
Our weighted cross-teaching loss, which integrates certainty measures to cross teach the transformer with a CNN network significantly improves the baseline method.
Masked domain learning trains the transformer with extra supervision signal, which significantly boosts the performance. 
The experiment result empirically confirms our previous analysis that \textbf{even more supervision signal is needed to effectively train the transformer}. 
The addition of adversarial masked domain adaptation to reduce the domain gap between the original and masked domain further improves the model performance. The experiment result reveals that \textbf{the domain gap between the original and masked domain can negatively affect semi-supervised learning performance and reducing the domain gap helps to boost semi-supervised learning performance}. Finally, the extension of adversarial masked image modeling to CNN network provides extra improvement.

\begin{figure}[t]
\centering
\includegraphics[width=0.8\textwidth]{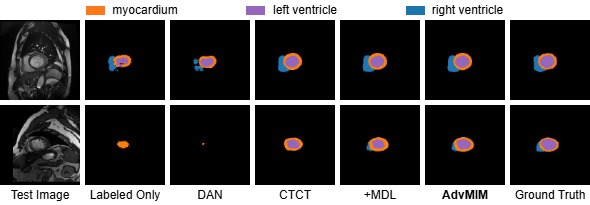}
\caption{Visual comparison with different methods on ACDC (3\%).} \label{fig:2}
\end{figure}

\begin{table}[t]
    \begin{minipage}[t]{.45\linewidth}
    \caption{Effect of Mask Ratio on ACDC (3\%) in Dice (\%) and HD (mm).
    } \label{table:3}
\begin{center}
\resizebox{0.8\linewidth}{!}{%
\begin{tabular}{c|ccccc}
    \hlineB{3}
    mask ratio&0.1&0.3&0.5&0.7&0.9 \\
    \hline
    Dice&84.7&85.0&85.1&\textbf{85.4}&80.5\\
    HD&2.4&2.3&2.1&\textbf{2.0}&3.8\\
    \hlineB{3}
\end{tabular}}
\end{center}
\label{plcpa_table3}
    \end{minipage}%
    \hfill
    \begin{minipage}[t]{.46\linewidth}
\caption{Sensitivity analysis of $\lambda_{adv}$ on ACDC (3\%) in Dice (\%) and HD (mm). 
} \label{table:4}
\begin{center}
\resizebox{0.7\linewidth}{!}{%
\begin{tabular}{c|ccccc}
    \hlineB{3}
    $\lambda_{adv}$&0.0001&0.001&0.01&0.1&1.0 \\
    \hline
    Dice&85.0&\textbf{85.4}&85.3&85.3&76.6\\
    HD&2.3&\textbf{2.0}&2.1&2.1&4.8\\
    \hlineB{3}
\end{tabular}}
\end{center}
    \end{minipage} 
\end{table}

\noindent\textbf{Comparison with SoTA Methods.} In Table~\ref{table:sota_comparison}, we present the comparison results of our method with state-of-the-art methods on different label partitions across three public datasets. As can be observed, for all label partitions and datasets, our method outperforms existing methods significantly.  For \textbf{ACDC}, our method outperforms the previous best method M-CnT by \textbf{10.1\%} in Dice and \textbf{8.7mm} in HD on 3\% partition, which is a tremendous improvement and highlights the effectiveness of our method in handling limited labeled data scenario.
For \textbf{Synapse}, our method significantly outperforms previous best method by \textbf{1.0\%} and \textbf{3.4\%} in Dice, \textbf{9.9mm} and \textbf{15.7mm} in HD on 15\% and 30\% partitions respectively. For \textbf{ISIC}, our method outperforms previous best method by \textbf{1.9\%} in Dice and \textbf{9.9mm} in HD on 3\% partition, where a tremendous improvement is observed in HD score.
We further observe that our method maintains strong performance even with a small label partition, where other methods fail. Specifically, our method achieves \textbf{85.4\%} and \textbf{79.8\%} in Dice for ACDC and ISIC with only 3\% labeled data, which highlights its remarkable applicability in real-world scenarios with extremely limited annotations.

\noindent\textbf{Visualization Results.} In Fig.~\ref{fig:2}, we present the visual comparison results of our method with different comparison methods. As shown in the figure, our method produces qualitatively much better segmentation masks when compared to existing adversarial training based method DAN~\cite{zhang2017deep}, state-of-the-art method CTCT~\cite{luo2022semi}, and our ablated masked domain learning method.

\noindent\textbf{Effectiveness of Mask Ratio.} In Table~\ref{table:3}, we present the effect of mask ratio on our method. Mask ratio controls the domain gap between masked and original domain. Too small mask ratio results in mask domain too similar to the original domain, which limits the extra supervision signal. Too large mask ratio increases the domain gap, which degrades the performance as supported by our Theorem~\ref{theorem1}. The default mask ratio of 0.7 gives the best performance.

\noindent\textbf{Sensitivity Analysis.} In Table~\ref{table:4}, we present the sensitivity analysis of our method on the hyperparameter $\lambda_{adv}$. Experiment results show that our method is robust to the change of $\lambda_{adv}$ in a wide range but too large value leads to poorer performance. The default value of $\lambda_{adv}=0.001$ gives the best performance.

\section{Conclusion}

In this paper, we propose an adversarial masked image modeling method to fully unleash the potential of transformer for semi-supervised medical image segmentation. Our key contributions include the construction of a masked domain with masked image modeling for effective training of transformer with extra supervision signal and a theoretical analysis showing that the domain gap between masked and original domain can negatively affect semi-supervised learning performance. Thus, we propose a novel adversarial masked domain adaptation loss to minimize the domain gap. We also extend adversarial masked image modeling to CNN network. Extensive experiments on three public medical image segmentation datasets demonstrate the effectiveness of our method, where our method outperforms existing methods significantly.

\noindent \textbf{Acknowledgement}
This work was supported by the Agency for Science, Technology, and Research (A*STAR) through its IEO Decentralised GAP Under Project I24D1AG085.

%
%
%
%
\bibliographystyle{splncs04}
\bibliography{main}
\end{document}